\documentclass[11pt,a4paper]{article}
\usepackage[final]{acl}
\usepackage{times}
\usepackage{latexsym}
\usepackage{booktabs}
\usepackage{graphics}
\usepackage{amsmath}
\usepackage{blindtext}
\usepackage{float}

\definecolor{forestgreen}{rgb}{0.22, 0.45, 0.19}
\definecolor{brickred}{rgb}{0.80, 0.20, 0.08}
\newcommand\gd{\textcolor{forestgreen}{$\big\uparrow$}}
\newcommand\bd{\textcolor{brickred}{$\big\downarrow$}}
\newcommand\nc{\textcolor{forestgreen}{\ \ \  }} 
\newcommand\OURS{KPT }
\usepackage{microtype}
\usepackage{graphicx}
\usepackage{multirow}
\usepackage[utf8]{inputenc}
\usepackage{amsfonts}
\usepackage{tcolorbox}
\usepackage{cleveref}
\usepackage[hang,flushmargin]{footmisc}
\crefname{section}{§}{§§}
\Crefname{section}{§}{§§}

\usepackage{booktabs,arydshln}

\makeatletter
\def\adl@drawiv#1#2#3{%
        \hskip.5\tabcolsep
        \xleaders#3{#2.5\@tempdimb #1{1}#2.5\@tempdimb}%
                #2\z@ plus1fil minus1fil\relax
        \hskip.5\tabcolsep}
\newcommand{\cdashlinelr}[1]{%
  \noalign{\vskip\aboverulesep
           \global\let\@dashdrawstore\adl@draw
           \global\let\adl@draw\adl@drawiv}
  \cdashline{#1}
  \noalign{\global\let\adl@draw\@dashdrawstore
           \vskip\belowrulesep}}
\makeatother

\newcommand{\smallcolor}[1]{
{\small\textcolor{forestgreen}{#1}}
}

\definecolor{mygray}{gray}{0.6}

\usepackage[T1]{fontenc}

\usepackage[utf8]{inputenc}

\title{Knowledgeable Prompt-tuning:\\Incorporating Knowledge into Prompt Verbalizer for Text Classification}

\author{Shengding Hu$^{1}$, Ning Ding$^{1}$, Huadong Wang$^{1*}$,  Zhiyuan Liu$^{1,2,4}$\thanks{\quad Corresponding authors: Z.Liu (liuzy@tsinghua.edu.cn), H.Wang (huadw2012@163.com)},\\
\textbf{Jingang Wang}$^{3}$\textbf{, Juanzi Li}$^{1}$\textbf{, Wei Wu}$^{3}$\textbf{, Maosong Sun}$^{1,2,4}$\\
\textsuperscript{1}Dept. of Comp. Sci. \& Tech., Institute for AI, Tsinghua University, Beijing, China\\
Beijing National Research Center for Information Science and Technology\\
\textsuperscript{2}Institute Guo Qiang, Tsinghua University, Beijing, China \textsuperscript{3}Meituan, Beijing, China\\
\textsuperscript{4}International Innovation Center of Tsinghua University, Shanghai, China\\
{\tt \{hsd20, dingn18\}@mails.tsinghua.edu.cn}
}

\begin{document}
\maketitle
\begin{abstract}

Tuning pre-trained language models (PLMs) with task-specific prompts has been a promising approach for text classification. Particularly, previous studies suggest that prompt-tuning has remarkable superiority in the low-data scenario over the generic fine-tuning methods with extra classifiers.  The core idea of prompt-tuning is to insert text pieces, i.e., template, to the input and transform a classification problem into a masked language modeling problem, where a crucial step is to construct a  projection, i.e., verbalizer, between a label space and a label word space. A verbalizer is usually handcrafted or searched by gradient descent, which may lack coverage and bring considerable bias and high variances to the results. In this work, we focus on incorporating external knowledge into the verbalizer, forming a \textit{knowledgeable prompt-tuning} (KPT), to improve and stabilize prompt-tuning.
Specifically, we expand the label word space of the verbalizer using external knowledge bases (KBs) and refine the expanded label word space with the PLM itself before predicting with the expanded label word space. Extensive experiments on zero and few-shot text classification tasks demonstrate the effectiveness of knowledgeable prompt-tuning. Our source code is publicly available at ~\url{https://github.com/thunlp/KnowledgeablePromptTuning}.



\end{abstract}

\section{Introduction}
\label{sec:intr}

Recent years have witnessed the prominence of Pre-trained Language Models (PLMs)~\cite{peters2018deep, radford2018improving, devlin2019bert,raffel2020exploring, xu2021pre} due to their superior performance on a wide range of language-related downstream tasks such as text classification~\cite{kowsari2019text}, question answering~\cite{rajpurkar2016squad}, and machine reading comprehension~\cite{nguyen2016ms}. To fathom the principles of such effectiveness of PLMs, researchers have conducted extensive studies and suggested that PLMs have obtained rich knowledge during pre-training~\cite{petroni2019language,davison-etal-2019-commonsense}. Hence, how to stimulate and exploit such knowledge is receiving increasing attention.


One conventional approach to achieve that is fine-tuning~\cite{devlin2019bert}, where we add extra classifiers on the top of PLMs and further train the models under classification objectives. Fine-tuning has achieved satisfying results on supervised tasks. However, since the extra classifier requires adequate training instances to tune, it is still challenging to apply fine-tuning in few-shot learning~\cite{brown2020language} and zero-shot learning~\cite{yin2019benchmarking} scenarios. Originated from GPT-3~\cite{brown2020language} and LAMA~\cite{petroni2019language, petroni2020how}, a series of studies using prompts~\cite{schick2020exploiting, liu2021gpt} for model tuning bridge the gap between pre-training objective and down-stream tasks, and demonstrate that such discrete or continuous prompts induce better performances for PLMs on few-shot and zero-shot tasks.



A typical way to use prompts is to wrap the input sentence into a natural language template and let the PLM conduct masked language modeling.
For instance, to classify the topic of a sentence $\mathbf{x}$: ``What's the relation between speed and acceleration?'' into the ``\textsc{Science}'' category, we wrap it into a template: ``A \texttt{[MASK]} question: $\mathbf{x}$''. The prediction is made based on the probability that the word ``science'' is filled in the ``\texttt{[MASK]}'' token. The mapping from \emph{label words} (e.g., ``science'' ) to the specific class (e.g.,  class \textsc{Science}) is called the \emph{verbalizer}~\cite{schick2020exploiting}, 
which bridges a projection between the vocabulary and the label space and has a great influence on the performance of classification~\cite{gao2020making}.  


 Most existing works use manual verbalizers~\cite{schick2020exploiting,schick2020s}, in which the designers manually think up a single word to indicate each class. To ease the human effort of designing the class name, some works propose to learn the label words using discrete search~\cite{schick2020automatically} or gradient descent~\cite{liu2021gpt, hambardzumyan-etal-2021-warp}.
However, the learned-from-scratch verbalizer, lack of human prior knowledge, is still \textit{considerably inferior} to the manual verbalizers (see Appendix~\ref{app:soft-pilot} for pilot experiments), especially in few-shot setting, and even not applicable in zero-shot setting, which leaves the manual verbalizer a decent choice in many cases.

However, manual verbalizers usually determine the predictions based on limited information. For instance, in the above example, the mapping \{\text{science}\}$\rightarrow $ \textsc{Science}   means that only predicting the word ``science'' for the \texttt{[MASK]} token is regarded as correct during inference, regardless of the predictions on other relevant words such as ``physics'' and ``maths'', which are also informative. 
 Such handcrafted one-one mapping limits the coverage of label words, thus lacking enough information for prediction and introducing bias into the verbalizer. Therefore, manual verbalizers are hard to be optimal in text classification, where the semantics of label words are crucial for predictions. 
 
 The optimization-based expansion, though can be combined with manual verbalizers to yield better performance, only induces a few words or embeddings that are close to the class name in terms of word sense or embedding distance. 
 Thus they are difficult to infer words across granularities (e.g. from ``science'' to ``physics'').  If we can expand the verbalizer of the above example into $\{\text{science, physics}\} \rightarrow \textsc{Science} $, the probability of making correct predictions will be considerably enhanced. 
 Therefore, to improve the coverage and reduce the bias of the manual verbalizer, we present to incorporate external knowledge into the verbalizers to facilitate prompt-tuning, namely, \emph{knowledgeable prompt-tuning} (KPT). Since our expansion is not based on optimization, it will also be more favorable for zero-shot learning.

Specifically, KPT contains three steps: construction, refinement, and utilization. (1) Firstly, in the construction stage,   we use external KBs to generate a set of label words for each label (in \cref{sec:cons}). Note that the expanded label words are not simply synonyms of each other, but cover different granularities and perspectives, thus are more comprehensive and unbiased than the class name. (2) Secondly, to cope with the noise in the unsupervised expansion of label words, we propose four refinement methods, namely, frequency refinement, relevance refinement, contextualized calibration, and learnable refinement (in \cref{sec:refine}), whose effectiveness is studied thoroughly in \cref{sec:exp}.
(3) Finally, we apply either a vanilla average loss function or a weighted average loss function for the utilization of expanded verbalizers, which map the scores on a set of label words to the scores of the labels.
 
 We conduct extensive experiments on zero-shot and few-shot text classification tasks. The empirical results show that KPT can reduce the error rate of classification by 16\%, 18\%, 10\%, 7\% on average in 0, 1, 5, 10 shot experiments, respectively, which
 shows the effectiveness of KPT. In addition to the performance boost, KPT also reduces the prediction variances consistently in few-shot experiments and yields more stable performances.

\begin{figure*}[!htbp]
    \centering
\scalebox{0.97}{
    \includegraphics[width = \linewidth]{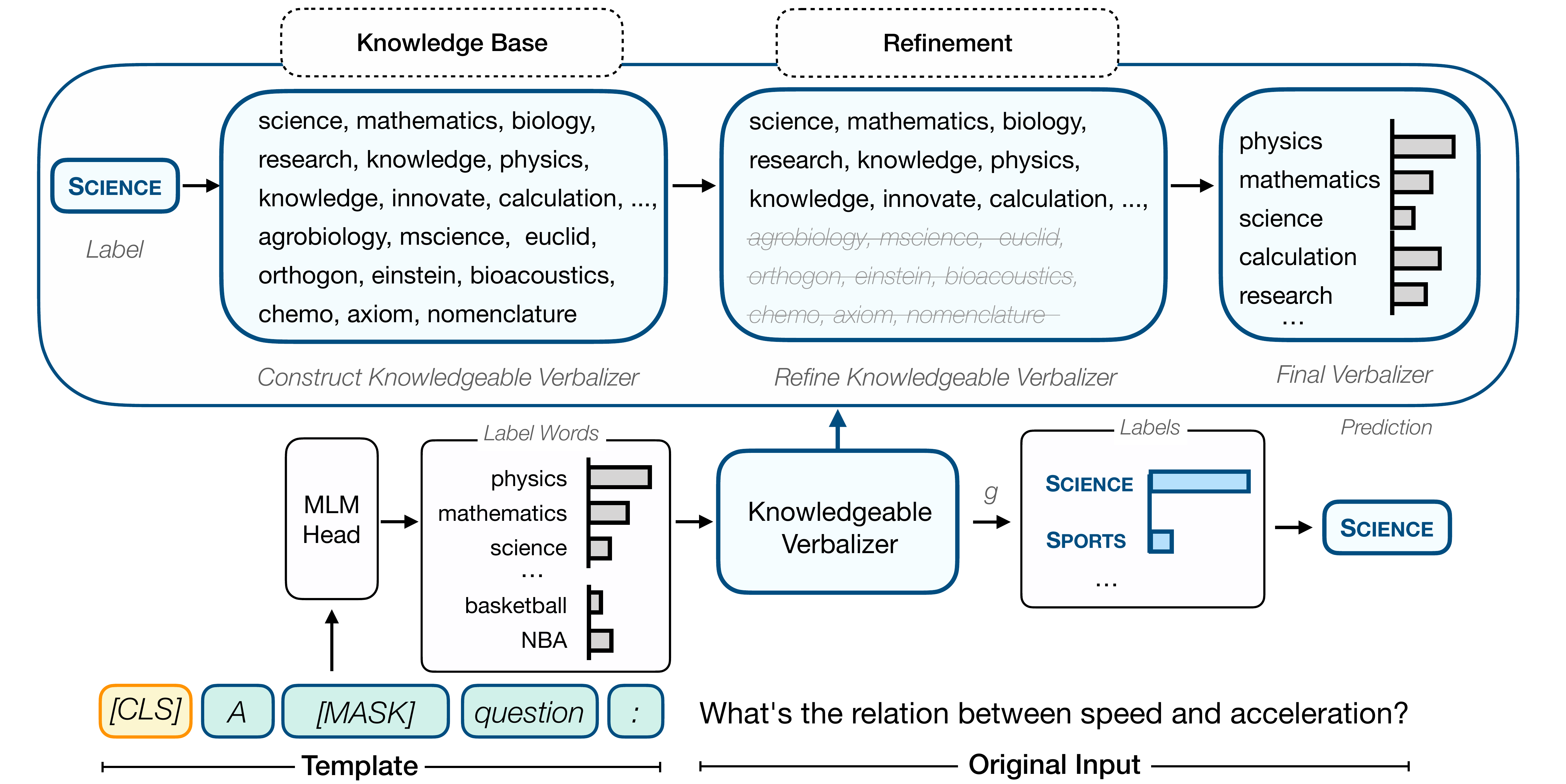}
    }
    \caption{The illustration of \OURS, the knowledgeable verbalizer maps the predictions over label words into labels. And the above part is the construction, refinement and utilization processes of \OURS.}
    \label{fig:my_label}
\end{figure*}

\section{Related Work}

Two groups of research are related to KPT: prompt-tuning, and the verbalizer construction.

\textbf{Prompt-tuning.} 
Since the emergence of  GPT-3~\cite{brown2020language}, prompt-tuning has received considerable attention. 
GPT-3~\cite{brown2020language} demonstrates that with prompt-tuning and in-context learning, the large-scale language models can achieve superior performance in the low-data regime. 
The following works~\cite{schick2020exploiting, schick2020s} argue that small-scale language models ~\cite{radford2018improving, devlin2019bert, liu2019roberta, lan2019albert} can also achieve decent performance using prompt-tuning. Prompt-tuning has been applied to a large variety of tasks such as Text Classification~\cite{schick2020exploiting}, Natural Language Understanding~\cite{schick2020s, liu2021gpt}
, Relation Extraction~\cite{han2021ptr,chen2021adaprompt}, and Knowledge Probing~\cite{petroni2019language, liu2021gpt}, etc.

\textbf{Verbalizer Construction.}
As introduced in \cref{sec:intr}, the verbalizer is an important component in prompt-tuning and has a strong influence on the performance of prompt-tuning~\cite{holtzman2021surface, gao2020making}. Most works use human-written verbalizers ~\cite{schick2020exploiting}, which are highly biased towards personal vocabulary and do not have enough coverage.  Some other studies~\cite{gao2020making, shin2020autoprompt,liu2021gpt,schick2020automatically} design automatic verbalizer searching methods for better verbalizer choices, however, their methods require adequate training set and validation set for optimization.  Moreover, the automatically determined verbalizers are usually synonym of the class name, which differs from our intuition of expanding the verbalizer with a set of diverse and comprehensive label words using external KB. 
~\citet{schick2020automatically} and~\citet{shin2020autoprompt} also try multiple label words for each class. 
The optimal size of their label words set for each class is generally less than 10, which lacks coverage when used in text classification tasks. 

\section{Knowledgeable Prompt-tuning}
\label{sec:method}
In this section, we present our methods to incorporate external knowledge into a prompt verbalizer. We first introduce the overall paradigm of prompt-tuning and then elucidate how to construct, refine and utilize the knowledgeable prompt.

\subsection{Overview}
Let $\mathcal{M}$ be a language model pre-trained on large scale corpora.
In text classification task, an input sequence $\mathbf{x} = (x_0,x_1,...,x_n)$ is classified into a class label $y\in \mathcal{Y}$. Prompt-tuning formalizes the classification task into a masked language modeling problem. Specifically, prompt-tuning wraps the input sequence with a \emph{template}, which is a piece of natural language text.
For example, assuming we need to classify the sentence $\mathbf{x}$ =``What's the relation between speed and acceleration?'' into label \textsc{Science} (labeled as 1) or \textsc{Sports} (labeled as 2), we wrap it into
\begin{equation*}
    \mathbf{x}_{\text{p}}=\text{\texttt{[CLS]} A \texttt{[MASK]} question : } \mathbf{x}
\end{equation*}
Then $\mathcal{M}$ gives the probability of each word $v$ in the vocabulary being filled in \texttt{[MASK]} token $P_\mathcal{M}(\texttt{[MASK]}=v|\mathbf{x}_{\text{p}})$.
To map the probabilities of words into the probabilities of labels, we define a \emph{verbalizer} as a mapping $f$ from a few words in the vocabulary, which form the \emph{label word} set $\mathcal{V}$, to the label space $\mathcal{Y}$, i.e., $f\colon \mathcal{V} \mapsto \mathcal{Y}$.  We use $\mathcal{V}_y$ to denote the subset of $\mathcal{V}$ that is mapped into a specific label $y$,  $\cup_{y\in\mathcal{Y}} \mathcal{V}_y = \mathcal{V}$. Then the probability of label $y$, i.e., $P(y|\mathbf{x}_{\text{p}})$, is calculated as
\begin{equation}
   P(y|\mathbf{x}_{\text{p}}) \!\!=\!\! g\left(P_\mathcal{M}(\texttt{[MASK]}\!\!\!=\!v|\mathbf{x}_{\text{p}})|v\in\mathcal{V}_y\right),
\end{equation}
where $g$ is a function transforming the probability of label words into the probability of the label.
 In the above example, regular prompt-tuning may define $\mathcal{V}_1=\{\text{``science''}\}$, $\mathcal{V}_2=\text{\{``sports''}\}$ and $g$ as an identity function, then if the probability of ``science'' is larger than ``sports'', we classify the instance into  \textsc{Science}. 

We propose KPT, which mainly focuses on using external knowledge to improve verbalizers in prompt-tuning.
In \OURS, we use KBs to generate multiple label words related to each class $y$, e.g., $\mathcal{V}_1 = \text{\{``science'',``physics'', ...\}}$. And we propose four refinement methods to eliminate the noise in the expanded $\mathcal{V}$. Finally, we explore the vanilla average and weighted average approaches for the utilization of the expanded $\mathcal{V}$.
The details are in the following sections.
\subsection{Verbalizer Construction}
\label{sec:cons}


The process of predicting masked words based on the context is not a single-choice procedure, that is, there is no standard correct answer, but abundant words may fit this context. 
Therefore, the label words mapped by a verbalizer should be equipped by two attributes: \textit{wide coverage} and \textit{little subjective bias}. Such a comprehensive projection is crucial to the imitation of pre-training, which is the essence of prompt-tuning. Fortunately, external structured knowledge could simultaneously meet both requirements. In this section, we introduce how we use external knowledge for two text classification tasks: topic classification and sentiment classification.

For topic classification, the core issue is to extract label words related to the topic from all aspects and granularities.  From this perspective, we choose Related Words~\footnote{\url{https://relatedwords.org}}, a knowledge graph $\mathcal{G}$ aggregated from multiple resources, including word embeddings, ConceptNet~\cite{speer2017conceptnet}, WordNet~\cite{pedersen2004wordnet}, etc., as our external KB. 
The edges denote "relevance" relations and are annotated with relevance scores. We presume the the name of each class $v_0$ is correct and use them as the anchor node to get the neighborhood nodes $N_{\mathcal{G}}(v_0)$ whose scores are larger than a threshold $\eta$ as the related words~\footnote{We take $\eta=0$ in the experiments}. Thus, each class is mapped into a set of label words $\mathcal{V}_{y} = N_{\mathcal{G}}(v_0) \cup \{v_0\}$.
For binary sentiment classification, the primary goal is to extend the binary sentiment to sentiment of more granualities and aspects. We use the sentiment dictionary summarized by previous researchers~\footnote{\url{https://www.enchantedlearning.com/wordlist/positivewords.shtml}}$^{,}$\footnote{ \url{https://www.enchantedlearning.com/wordlist/negativewords.shtml}}.
Several examples of the label words in the \OURS are in Table~\ref{tab:label_words_examples}.



\begin{table*}[!htbp]
    \centering
\scalebox{0.90}{
    \begin{tabular}{ccl}
    \toprule
  Dataset &  Label & \multicolumn{1}{c}{Label Words} \\
    \midrule
  \multirow{2}{*}{AG's News}  &  \textsc{Politics} &  politics, government, diplomatic, law, aristotle, diplomatical, governance ...\\
    &  \textsc{Sports}   & sports, athletics, gymnastics, sportsman, competition, cycling, soccer ... \\
      \midrule
 \multirow{2}{*}{IMDB} &    \textsc{Negative} & abysmal, adverse, alarming, angry, annoy, anxious, apathy, appalling ...\\
    &  \textsc{Positive} & absolutely, accepted, acclaimed, accomplish, accomplishment  ...\\
      \bottomrule
    \end{tabular}
}
    \caption{Examples of the expanded label words.}
    \label{tab:label_words_examples}
\end{table*}

\subsection{Verbalizer Refinement}
\label{sec:refine}

Although we have constructed a knowledgeable verbalizer that contains comprehensive label words, the collected label words can be very noisy since the vocabulary of the KB is not tailored for the PLM. Thus it is necessary to refine such verbalizer by retaining high-quality words. In this section, we propose four refinement methods addressing different problems of the noisy label words.



\textbf{Frequency Refinement.} The first problem is to handle the rare words. We assume that several words in the KB are rare to the PLM, thus the prediction probabilities on these words tend to be inaccurate.  Instead of using a word-frequency dictionary, we propose to use \emph{contextualized prior} of the label words to remove these words. Specifically, given a text classification task, we denote the distribution of the sentences $\mathbf{x}$ in the corpus as $\mathcal{D}$. For each sentence in the distribution, we wrap it into the template and calculate the predicted probability for each label word $v$ in the masked position $P_\mathcal{M}(\texttt{[MASK]}\!\!\!=\!v|\mathbf{x}_{\text{p}})$. By taking the expectation of the probability over the entire distribution of sentences, we can get the prior distribution of the label words in the masked position. We formalize it as
\begin{equation}
 P_{\mathcal{D}}(v)\! =\! \mathbb{E}_{\mathbf{x}\sim \mathcal{D}} P_\mathcal{M}(\texttt{[MASK]}\!\!\!=\!v|\mathbf{x}_{\text{p}}).
\end{equation}
Empirically, we found that using a small-size \emph{unlabeled support set} $\tilde{\mathcal{C}}$ sampled from the training set and with labels removed, will yield a satisfying estimate of the above expectation.  Thus, assuming that the input samples $\{\mathbf{x}\in \tilde{\mathcal{C}}\}$ have a uniform prior distribution, the contextualized prior is approximated by
\begin{equation}
    P_{\mathcal{D}}(v) \approx \frac{1}{|\tilde{\mathcal{C}}|} \sum_{\mathbf{x}\in \tilde{\mathcal{C}}} P_\mathcal{M}(\texttt{[MASK]}\!\!\!=\!v|\mathbf{x}_{\text{p}}).
\end{equation}
Then we remove the label words whose prior probabilities are less than a threshold. Details can be found in Appendix~\ref{App: filter}.

\textbf{Relevance Refinement.}
As our construction of knowledgeable label words is fully unsupervised, some label words may be more relevant to their belonging class than the others. 
To measure the relevance of a label word to each class, we obtain the prediction probability of the label word on the support set $\tilde{\mathcal{C}}$ as the vector representation $\mathbf{q}^{v}$ of the label words, i.e.,  $\mathbf{q}^{v}$'s $i$-th element is
\begin{equation}
    \mathbf{q}^{v}_i = P_\mathcal{M}([\texttt{MASK}]=v|\mathbf{x}_{i{\text{p}}}), \mathbf{x}_i\in \tilde{\mathcal{C}},
\end{equation}
where $\mathbf{x}_{i{\text{p}}}$ represents the sentence $x_i$ combined with the template $\text{p}$.

To estimate the class's representation, we presume that the name of each class $v_0$, such as ``science'' for \textsc{Science}, though lack of coverage, is very relevant to the class. Then we use the vector representation $\mathbf{q}^{v_0}$ of the these names as the class's representation $\mathbf{q}^{y}$. Therefore the relevance score between a label word $v$ and a class $y$ is calculated as the cosine similarity between the two representation:
\begin{equation}
    r(v, y) = \operatorname{cos}(\mathbf{q}^{v}, \mathbf{q}^{y}) = \operatorname{cos}(\mathbf{q}^{v}, \mathbf{q}^{v_0}).
\end{equation}

Moreover, some label words may contribute positively to multiple classes, resulting in confusion between classes. For example, the potential label word ``physiology'' of class \textsc{Science} may also be assigned with a high probability in a sentence of class \textsc{Sports}. To mitigate such confusion and filter the less relevant label words, we design a metric that favors the label word with high relevance \emph{merely} to its belonging class and low relevance to other classes:
\begin{equation}
    R(v) = r(v, f(v)) \frac{|\mathcal{Y}|-1}{\sum_{y\in \mathcal{Y}, y\neq f(v)}(r(v,y))},
\label{equ: rr_tfidf}
\end{equation}
where $f(v)$ is the corresponding class of $v$.

Ideally, a good label word should at least has a higher relevance score for its belonging class than the average relevance score for the other classes. Therefore, we remove the label words with $R(v)<1$. In practice, we have a slight modification to Equation~\eqref{equ: rr_tfidf}, 
please refer to appendix~\ref{App: filter} for details.

Essentially, this Relevance Refinement adopts the idea of  the classical TF-IDF~\cite{jones1972statistical} algorithm which estimates the relevance of a word to a document. It prefers to use a word that is relevant to a specific document while irrelevant to other documents as the keyword of the document. In KPT, a class is analogous to a document, while a label word is comparable to the word in the document. From this perspective,  equation~\eqref{equ: rr_tfidf} is a variant of TF-IDF metric. 

\textbf{Contextualized Calibration.}\quad
\label{sec: refine-CC}
The third problem is the drastic difference in the prior probabilities of label words. As previous works~\cite{zhao2021calibrate, holtzman2021surface} have shown, some label words are less likely to be predicted than the others, regardless of the label of input sentences, resulting in a biased prediction. In our setting, the label words in the KB tend to have more diverse prior probabilities, resulting in a severer problem (see Table~\ref{tab:experiment-zero-shot}). Therefore,  we use the contextualized prior of label words to calibrate the predicted distribution, namely, \emph{contextualized calibration} (CC):
\begin{equation}
    \tilde{P}_\mathcal{M}(\!\texttt{[MASK]}\!\!\!=\!v|\mathbf{x}_{\text{p}}) \!\propto\! \frac{P_\mathcal{M}(\!\texttt{[MASK]}\!\!\!=\!v|\mathbf{x}_{\text{p}})}{P_{\mathcal{D}}(v)}
\end{equation}
where $P_{\mathcal{D}}(v)$ is the prior probability of the label word. The final probability is normalized to 1.

\textbf{Learnable Refinement}. In few-shot learning, the refinement can be strengthen by a learning process. Specifically  we assign a learnable weight $w_v$ to each label word $v$ (may be already refined by the previous methods). The weights form a vector $\mathbf{w}\in \mathbb{R}^{|\mathcal{V}|}$, which is initialized to be a zero vector. The  weights are normalized within each $\mathcal{V}_y$:

\begin{equation}
    \alpha_v = \frac{\operatorname{exp}(w_v)}{\sum_{u\in \mathcal{V}_y} \operatorname{exp}(w_u)}.
\end{equation}

 Intuitively, in the training process, a small weight is expected to be learned for a noisy label word to minimize its influence on the prediction. Note that in few-shot setting,  calibration may not be necessary because the probability of a label word can be trained to the desired magnitude, i.e., $\tilde{P}_\mathcal{M}(\texttt{[MASK]}\!\!\!=\!v|\mathbf{x}_{\text{p}}) ={P_\mathcal{M}(\texttt{[MASK]}\!\!\!=\!v|\mathbf{x}_{\text{p}})}$. 


In addition to these refinement methods, since many label words are out-of-vocabulary for the PLM and are split into multiple tokens by the tokenizer. For these words, we simply use the average prediction score of each token as the prediction score for the word. The influence of this simple approach is studied in Appendix~\ref{app:experiment-st}.

\subsection{Verbalizer Utilization}
\label{sec:use}

The final problem is how to map the predicted probability on each refined label word to the decision of the class label $y$.


 \textbf{Average.} After refinement, we can assume that each label word of a class contributes equally to predicting the label. Therefore, we use the average of the predicted scores on $\mathcal{V}_y$ as the predicted  score for label $y$. 
The predicted label $\hat{y}$ is

\begin{equation}
\hat{y}\!=\!
    \mathrm{arg}\!\max_{y\in\mathcal{Y}}\frac{\sum_{v \in \mathcal{V}_{y}}\! \tilde{P}_\mathcal{M}(\!\texttt{[MASK]}=v|\mathbf{x}_{\text{p}})}{|\mathcal{V}_y|}.
\end{equation}

We use this method in zero-shot learning since there is no parameter to be trained.

\textbf{Weighted Average.} In few-shot setting, supported by  the Learnable Refinement,  we adopt a weighted average of label words' scores as the prediction score. The refinement weights $\alpha_v$ are used as the weights for averaging. Thus, the predicted $\hat{y}$ is
\begin{equation}
\hat{y} \!= \operatorname{argmax}_{y\in\mathcal{Y}} \frac{\exp\left({s(y|\mathbf{x}_{\text{p}})}\right)}{\sum_{y'}\exp\left({s({y'}|\mathbf{x}_{\text{p}})}\right)},
\end{equation}
where $s(y|\mathbf{x}_{\text{p}})$ is
\begin{equation}
        s(y|\mathbf{x}_{\text{p}})\!=\!\! \sum_{v\in\mathcal{V}_y} \!\!\alpha_v  \operatorname{log}{P}_\mathcal{M}(\texttt{[MASK]}\!\!\!=\!v|\mathbf{x}_{\text{p}}).
\end{equation}
This objective function is suitable for continuous optimization by applying a cross-entropy loss on the predicted probability.

\subsection{Theoretical Illustration of \OURS}
We provide a theoretical illustration of the KPT framwork in Appendix~\ref{theoretical}.


\begin{table*}[!htbp]
\begin{center}
\scalebox{0.85}{
\begin{tabular}{lccccc}
\toprule
 Method & AG's News   & DBPedia & Yahoo& Amazon & IMDB  \\
\midrule
PT & 75.1 $\pm$ 6.2 \smallcolor{(79.0)} & 66.6 $\pm$ 2.3 \smallcolor{(68.4)} & 45.4 $\pm$ 7.0 \smallcolor{(52.0)} & 80.2 $\pm$ 8.8 \smallcolor{(87.8)} & 86.4 $\pm$ 4.0 \smallcolor{(92.0)} \\
PT+CC & 79.9 $\pm$ 0.7 \smallcolor{(81.0)} & 73.9 $\pm$ 4.9 \smallcolor{(82.6)} & 58.0 $\pm$ 1.4 \smallcolor{(58.8)} & 91.4 $\pm$ 1.6 \smallcolor{(93.5)} & 91.6 $\pm$ 3.0 \smallcolor{(93.7)} \\

\midrule
\OURS & \textbf{84.8 $\pm$ 1.2} \smallcolor{\smallcolor{(\textbf{86.7})}} & \textbf{82.2 $\pm$ 5.4} \smallcolor{(\textbf{87.4})} & \textbf{61.6 $\pm$ 2.2} \smallcolor{(\textbf{63.8})} & \textbf{92.8 $\pm$ 1.2} \smallcolor{(\textbf{94.6})} & \textbf{91.6 $\pm$ 2.7} \smallcolor{(94.0)} \\
 \cdashlinelr{2-6}
\ \ -FR & 82.7 $\pm$ 1.5 \smallcolor{(85.0)} & 81.8 $\pm$ 4.6 \smallcolor{(86.2)} & 60.9 $\pm$ 1.5 \smallcolor{(62.7)} & \textbf{92.8 $\pm$ 1.2} \smallcolor{(\textbf{94.6})} & 91.6 $\pm$ 2.8 \smallcolor{(\textbf{94.1})} \\
\ \ \ \ -RR & 81.4 $\pm$ 1.5 \smallcolor{(83.7)} & 81.4 $\pm$ 4.5 \smallcolor{(85.8)} & 60.1 $\pm$ 1.0 \smallcolor{(61.4)} & \textbf{92.8 $\pm$ 1.2} \smallcolor{(\textbf{94.6})} & 91.6 $\pm$ 2.8 \smallcolor{(\textbf{94.1})} \\
\ \ \ \ \ \ -CC & 55.5 $\pm$ 2.8 \smallcolor{(58.3)} & 64.5 $\pm$ 6.8 \smallcolor{(73.0)} & 42.4 $\pm$ 5.0 \smallcolor{(46.8)} &
86.2 $\pm$ 5.7 \smallcolor{(92.5)} & 90.3 $\pm$ 2.8 \smallcolor{(\textbf{94.1})} \\
  
\bottomrule
\end{tabular}}
\end{center}
\caption{Results of zero-shot text classification.
The results of the best templates are shown in the brackets. 
Indentation means that the experimental configuration is a modification based on the up-level indentation.}
\label{tab:experiment-zero-shot}
\end{table*}

\section{Experiments}
\label{sec:exp}
We evaluate \OURS on five text classification datasets to demonstrate the effectiveness of incorporating external knowledge into prompt-tuning. 

\subsection{Datasets and Templates}

We carry out experiments on three topic classification datasets: AG's News~\cite{zhang2015character}, DBPedia~\cite{lehmann2015dbpedia}, and Yahoo~\cite{zhang2015character}, and two sentiment classification datasets: IMDB~\cite{maas2011learning} and Amazon~\cite{mcauley2013hidden}. The statistics of the datasets are shown in Table~\ref{tab:dataset_stat}. The detailed information and the statistics of each dataset is in Appendix~\ref{App:dataset-template}. 

We test all prompt-based methods using four manual templates and report both the average results (with standard error) of the four templates and the results of the best template (shown in\smallcolor{(brackets)}).
The reasons for using manual templates and the specific templates for each dataset are in Appendix~\ref{App:dataset-template}.
\subsection{Experiment Settings}
Our experiments are based on OpenPrompt~\cite{ding2021openprompt}, which is an open-source toolkit to conduct prompt learning. 
For the PLM, we use $\text{RoBERTa}_\text{large}$~\cite{liu2019roberta}
for all experiments. For test metrics, we use Micro-F1 in all experiments.
For all zero-shot experiments, we repeat the experiments 3 times using different random seeds if randomness is introduced in the experiments, and for all few-shot experiments, we repeat 5 times. Note that considering the four templates and five/three random seeds, each reported score of prompt-based methods is \textit{the average of 20/12 experiments}, which greatly reduces the randomness of the evaluation results. 
For the refinement based on the support set $\tilde{\mathcal{C}}$, the size of the unlabeled support set $|\tilde{\mathcal{C}}|$ is 200.  
For few-shot learning, we conduct 1, 5, 10, and 20-shot experiments. For a $k$-shot experiment, we sample $k$ instances of each class from the original training set to form the few-shot training set and sample another $k$ instances per class to form the validation set. We tune the entire model for 5 epochs and choose the checkpoint with the best validation performance to test. 
Other hyper-parameters 
can be found in Appendix~\ref{app:exp_settings}.


\begin{table*}[!ht]
\begin{center}
\scalebox{0.79}{
\begin{tabular}{lllllll}
\toprule
Shot & Method  & AG's News   & DBPedia & Yahoo & Amazon & IMDB  \\
\midrule
\multirow{8}{*}{1} & FT & 19.8 $\pm$ 10.4 & 8.6 $\pm$ 4.5 &11.1 $\pm$ 4.0 & 49.9 $\pm$ 0.2 & 50.0 $\pm$ 0.0  \\
 & PT & 80.0 $\pm$ 6.0 \smallcolor{\smallcolor{(84.4)}} & 92.2 $\pm$ 2.5 \smallcolor{\smallcolor{(94.3)}} & 54.2 $\pm$ 3.1 \smallcolor{\smallcolor{(55.7)}} & 91.9 $\pm$ 2.7 \smallcolor{\smallcolor{(93.2)}} & 91.2 $\pm$ 3.7 \smallcolor{\smallcolor{(93.7)}} \\
& AUTO & 52.8 $\pm$ 9.8 \smallcolor{\smallcolor{(57.6)}} & 63.0 $\pm$ 8.9 \smallcolor{\smallcolor{(68.3)}} & 23.3 $\pm$ 4.5 \smallcolor{\smallcolor{(25.0)}} & 66.6 $\pm$ 12.5 \smallcolor{\smallcolor{(72.7)}} & 75.5 $\pm$ 15.5 \smallcolor{\smallcolor{(83.1)}} \\
& SOFT & 80.0 $\pm$ 5.6 \smallcolor{\smallcolor{(82.4)}} & 92.3 $\pm$ 2.3 \smallcolor{\smallcolor{(93.3)}} & 54.3 $\pm$ 2.7 \smallcolor{\smallcolor{(55.9)}} & 90.9 $\pm$ 5.8 \smallcolor{\smallcolor{(93.6)}} & 89.4 $\pm$ 8.9 \smallcolor{\smallcolor{(93.1)}} \\
 \cmidrule{2-7}
& \OURS & \textbf{83.7 $\pm$ 3.5} \smallcolor{\smallcolor{(\textbf{84.6})}} & \textbf{93.7 $\pm$ 1.8} \smallcolor{\smallcolor{(\textbf{95.3})}} & \textbf{63.2 $\pm$ 2.5} \smallcolor{\smallcolor{(\textbf{64.1})}} & 93.2 $\pm$ 1.3 \smallcolor{\smallcolor{(\textbf{93.9})}} & 92.2 $\pm$ 3.0 \smallcolor{\smallcolor{(\textbf{93.6})}} \\
\cdashlinelr{2-7}
& \ \ - LR & 83.5 $\pm$ 3.8 \smallcolor{\smallcolor{(84.3)}} & 93.0 $\pm$ 1.8 \smallcolor{\smallcolor{(94.5)}} & 62.2 $\pm$ 2.9 \smallcolor{\smallcolor{(63.6)}} & \textbf{93.3 $\pm$ 1.3} \smallcolor{\smallcolor{(\textbf{93.9})}} & 92.2 $\pm$ 2.8 \smallcolor{\smallcolor{(\textbf{93.6})}} \\
& \ \ - RR & 82.2 $\pm$ 3.2 \smallcolor{\smallcolor{(82.6)}} & 92.9 $\pm$ 1.8 \smallcolor{\smallcolor{(94.1)}} & 61.3 $\pm$ 4.2 \smallcolor{\smallcolor{(62.5)}} & 93.1 $\pm$ 1.5 \smallcolor{\smallcolor{(93.7)}} & \textbf{92.6 $\pm$ 1.7} \smallcolor{\smallcolor{(\textbf{93.6})}} \\
& \ \ - RR - LR & 81.8 $\pm$ 3.3 \smallcolor{\smallcolor{(82.5)}} & 91.3 $\pm$ 1.7 \smallcolor{\smallcolor{(92.6)}} & 60.7 $\pm$ 4.2 \smallcolor{\smallcolor{(61.4)}} & 93.2 $\pm$ 1.5 \smallcolor{\smallcolor{(93.9)}} & 92.6 $\pm$ 1.5 \smallcolor{\smallcolor{(93.5)}} \\
\midrule
\multirow{8}{*}{5} & FT &  37.9 $\pm$ 10.0  & 95.8 $\pm$ 1.3  & 25.3 $\pm$ 14.2 & 52.1 $\pm$ 1.3     &  51.4 $\pm$ 1.4 \\
 & PT & 82.7 $\pm$ 2.7 \smallcolor{\smallcolor{(84.0)}} & 97.0 $\pm$ 0.6 \smallcolor{\smallcolor{(\textbf{97.3})}} & 62.4 $\pm$ 1.7 \smallcolor{\smallcolor{(63.9)}}  & 92.2 $\pm$ 3.3 \smallcolor{\smallcolor{(93.5)}} & 91.9 $\pm$ 3.1 \smallcolor{\smallcolor{(92.7)}} \\
& AUTO &72.2 $\pm$ 10.1 \smallcolor{\smallcolor{(75.6)}} & 88.8 $\pm$ 3.9 \smallcolor{\smallcolor{(91.5)}} & 49.6 $\pm$ 4.3 \smallcolor{(51.2)} & 87.5 $\pm$ 7.4 \smallcolor{(90.8)} & 86.8 $\pm$ 10.1 \smallcolor{(92.1)} \\
& SOFT &82.8 $\pm$ 2.7 \smallcolor{(84.3)} & 97.0 $\pm$ 0.6 \smallcolor{(97.2)} & 61.8 $\pm$ 1.8 \smallcolor{(63.1)} & 93.2 $\pm$ 1.6 \smallcolor{(\textbf{94.2})} & 91.6 $\pm$ 3.4 \smallcolor{(\textbf{93.9})} \\
 \cmidrule{2-7}
& \OURS & 85.0 $\pm$ 1.2 \smallcolor{(\textbf{85.9})} & 97.1 $\pm$ 0.4 \smallcolor{(97.3)} & \textbf{67.2 $\pm$ 0.8} \smallcolor{(\textbf{67.8})} & 93.4 $\pm$ 1.9 \smallcolor{(94.1)} & 92.7 $\pm$ 1.5 \smallcolor{(92.9)} \\
\cdashlinelr{2-7}
&\ \ - LR & \textbf{85.1 $\pm$ 1.0} \smallcolor{(85.8)} & 97.1 $\pm$ 0.4 \smallcolor{(97.2)} & 67.0 $\pm$ 1.1 \smallcolor{(67.5)} & 93.4 $\pm$ 1.9 \smallcolor{(94.1)} & 92.8 $\pm$ 1.5 \smallcolor{(93.0)} \\
&\ \ - RR & 84.3 $\pm$ 1.8 \smallcolor{(84.9)} & \textbf{97.2 $\pm$ 0.4} \smallcolor{(\textbf{97.3})} & \textbf{67.2 $\pm$ 0.8} \smallcolor{(67.7)} & \textbf{93.6 $\pm$ 1.4} \smallcolor{(94.1)} & \textbf{93.0 $\pm$ 2.0} \smallcolor{(93.8)} \\
& \ \ - RR - LR & 84.2 $\pm$ 1.7 \smallcolor{(84.5)} & 97.1 $\pm$ 0.4 \smallcolor{(97.3)} & 66.6 $\pm$ 1.4 \smallcolor{(67.5)} & 93.4 $\pm$ 2.0 \smallcolor{(94.1)} & 93.0 $\pm$ 2.1 \smallcolor{(93.8)} \\
\midrule
\multirow{8}{*}{10} & FT & 75.9 $\pm$ 8.4   & 93.8 $\pm$ 2.2 & 43.8 $\pm$ 17.9 & 83.0 $\pm$ 7.0    & 76.2 $\pm$ 8.7  \\
 & PT & 84.9 $\pm$ 2.4 \smallcolor{(86.1)} & 97.6 $\pm$ 0.4 \smallcolor{(97.8)} & 64.3 $\pm$ 2.2 \smallcolor{(64.8)} & 93.9 $\pm$ 1.3 \smallcolor{(94.6)} & \textbf{93.0 $\pm$ 1.7} \smallcolor{(\textbf{94.0})} \\
 & AUTO & 81.4 $\pm$ 3.8 \smallcolor{(84.1)} & 91.5 $\pm$ 3.4 \smallcolor{(95.1)} & 58.7 $\pm$ 3.1 \smallcolor{(60.9)} & 93.7 $\pm$ 1.2 \smallcolor{(94.5)} & 91.1 $\pm$ 5.1 \smallcolor{(93.3)} \\
 & SOFT &85.0 $\pm$ 2.8 \smallcolor{(86.7)} & 97.6 $\pm$ 0.4 \smallcolor{(97.8)} & 64.5 $\pm$ 2.2 \smallcolor{(65.0)} & 93.9 $\pm$ 1.7 \smallcolor{(93.9)} & 91.8 $\pm$ 2.6 \smallcolor{(93.0)} \\
 \cmidrule{2-7}
& \OURS & \textbf{86.3 $\pm$ 1.6} \smallcolor{(87.0)} & \textbf{98.0 $\pm$ 0.2} \smallcolor{(\textbf{98.1})} & \textbf{68.0 $\pm$ 0.6} \smallcolor{(\textbf{68.2})} & 93.8 $\pm$ 1.2 \smallcolor{(94.1)} & 92.9 $\pm$ 1.8 \smallcolor{(93.3)} \\
\cdashlinelr{2-7}
& \ \ - LR & 85.9 $\pm$ 1.9 \smallcolor{(\textbf{87.1})} & 98.0 $\pm$ 0.2 \smallcolor{(\textbf{98.1})} & 67.9 $\pm$ 0.7 \smallcolor{(\textbf{68.2})} & 93.9 $\pm$ 1.1 \smallcolor{(94.1)} & \textbf{93.0 $\pm$ 1.7} \smallcolor{(93.2)} \\
& \ \ - RR & 85.6 $\pm$ 1.4 \smallcolor{(86.2)} & 97.9 $\pm$ 0.2 \smallcolor{(98.0)} & 67.5 $\pm$ 1.1 \smallcolor{(68.1)} & {94.0 $\pm$ 1.0} \smallcolor{(\textbf{94.7})} & 92.7 $\pm$ 2.1 \smallcolor{(93.0)} \\
& \ \ - RR - LR & 85.1 $\pm$ 1.4 \smallcolor{(86.0)} & 97.8 $\pm$ 0.2 \smallcolor{(97.8)} & 66.8 $\pm$ 1.1 \smallcolor{(67.6)} & \textbf{94.1 $\pm$ 0.9} \smallcolor{(94.8)} & 93.0 $\pm$ 2.0 \smallcolor{(93.4)} \\
\midrule
\multirow{8}{*}{20} & FT & 85.4 $\pm$ 1.8     & 97.9 $\pm$ 0.2  & 54.2 $\pm$ 18.1 & 71.4 $\pm$ 4.3   & 78.5 $\pm$ 10.1   \\
 & PT & 86.5 $\pm$ 1.6 \smallcolor{(87.0)} & 97.9 $\pm$ 0.3 \smallcolor{(98.1)} & 67.2 $\pm$ 1.1 \smallcolor{(67.5)} & 93.5 $\pm$ 1.0 \smallcolor{(94.4)} & 93.0 $\pm$ 1.1 \smallcolor{(93.6)} \\
& AUTO &85.7 $\pm$ 1.4 \smallcolor{(86.1)} & 92.2 $\pm$ 2.7 \smallcolor{(94.9)} & 65.0 $\pm$ 1.8 \smallcolor{(66.9)} & \textbf{93.9 $\pm$ 1.1} \smallcolor{(94.1)} & 92.8 $\pm$ 2.0 \smallcolor{(\textbf{94.0})} \\
&SOFT &86.4 $\pm$ 1.7 \smallcolor{(87.1)} & 98.0 $\pm$ 0.3 \smallcolor{(98.1)} & 67.4 $\pm$ 0.7 \smallcolor{(67.5)} & 93.8 $\pm$ 1.6 \smallcolor{(94.2)} & \textbf{93.5 $\pm$ 0.9} \smallcolor{(\textbf{94.0})} \\
 \cmidrule{2-7}
 & \OURS &87.2 $\pm$ 0.8 \smallcolor{(87.5)} & \textbf{98.1 $\pm$ 0.3} \smallcolor{(\textbf{98.2})} & 68.9 $\pm$ 0.8 \smallcolor{(69.3)} & 93.7 $\pm$ 1.6 \smallcolor{(94.4)} & 93.1 $\pm$ 1.1 \smallcolor{(93.5)} \\
 \cdashlinelr{2-7}
 & \ \ - LR & \textbf{87.7 $\pm$ 0.6} \smallcolor{(\textbf{87.8})} & \textbf{98.1 $\pm$ 0.3} \smallcolor{(\textbf{98.2})} & 68.8 $\pm$ 0.9 \smallcolor{(\textbf{69.8})} & 93.4 $\pm$ 2.3 \smallcolor{(94.3)} & 93.4 $\pm$ 0.9 \smallcolor{(93.6)} \\
& \ \ - RR &87.3 $\pm$ 0.8 \smallcolor{(87.5)} & \textbf{98.1 $\pm$ 0.3} \smallcolor{(\textbf{98.2})} & 68.8 $\pm$ 0.9 \smallcolor{(68.9)} & 93.6 $\pm$ 1.3 \smallcolor{(94.2)} & 93.1 $\pm$ 0.8 \smallcolor{(93.6)} \\
& \ \ - RR - LR & 87.1 $\pm$ 0.9 \smallcolor{(87.4)} & \textbf{98.1 $\pm$ 0.3} \smallcolor{(\textbf{98.2})} & \textbf{69.0 $\pm$ 0.7} \smallcolor{(69.3)} & 93.7 $\pm$ 0.9 \smallcolor{(\textbf{94.5})} & 93.1 $\pm$ 0.8 \smallcolor{(93.7)} \\
\bottomrule
\end{tabular}}
\end{center}
\caption{Results of 1/5/10/20-shot text classification. 
Indentation means that the experimental configuration is a modification based on the up-level indentation.
}
\label{tab:experiment-few-shot}
\end{table*}

\subsection{Baselines}
In this subsection, we introduce the baselines we compare with.
To better understand our proposed methods, we also compare within the performance of \OURS using different configuration. 

\textbf{Fine-tuning (FT).}\quad Traditional fine-tuning method inputs the hidden embedding of \texttt{[CLS]} token of the PLM into the classification layer to make predictions. Note that fine-tuning can not be applied to the zero-shot setting, since the classification layer is randomly initialized. 


\textbf{Prompt-tuning (PT).}\quad  The regular prompt-tuning method uses the class name as the only label word for each class, which is used in PET~\cite{schick2020exploiting} and most existing works. 
For a fair comparison, we do not use the tricks in PET, such as self-training and prompt ensemble, which are orthogonal to our contributions.

\textbf{Automatic Verbalizer (AUTO).}\quad The automatic verbalizer is proposed by  PETAL~\cite{schick2020automatically}, which uses \emph{labeled} data to select the most informative label words \emph{inside} a PLM's vocabulary. It is targeted at the situation when no manually defined class names are available. It's not obvious how to combine it with the manually defined class name to boost the performance, and how it can be applied in a zero-shot setting. Therefore we only compare it in the few-shot setting with no class name information given. 

\textbf{Soft Verbalizer (SOFT).}\quad The soft verbalizer is proposed by WARP~\cite{hambardzumyan-etal-2021-warp}. They use a continuous vector for each class and use the dot product between the masked language model output and the class vector to produce the probability for each class. 
In our experiments, its class vectors are initialized with the class names' word embedding, since it is more effective with manual class names as the initial values (see Appendix~\ref{app:soft-pilot}). As an optimization-based method, Soft Verbalizer is not applicable in the zero-shot setting. 

\textbf{PT+CC.} For zero-shot setting, we further introduce PT combined with our proposed contextualized calibration\footnote{The same support sets are used as \OURS.} as a baseline 
to see how much improvement is made by contextualized calibration instead of knowledgeable verbalizers. 

For \textbf{\OURS}, we experiment with different variants to better understand the proposed methods such as refinement. 
\textbf{-FR}, \textbf{-RR}, \textbf{-CC} and \textbf{-LR} is the variant that does not conduct Frequency Refinement,  Relevance Refinement,  Contextualized Calibration, and Learnable Refinement, respectively.
In few-shot experiments, we presume that the supervised training data can train the output probability of each label word to the desired magnitude, thus we don't use CC and FR in the \OURS. This decision is justified in Appendix~\ref{app:fewshot-cali}.

\subsection{Main Results}
In this subsection, we introduce the specific results and provide possible insights of \OURS.

\textbf{Zero-shot.} \quad From Table~\ref{tab:experiment-zero-shot}, we see that all the variants of \OURS, except for KPT-CC, consistently outperforms PT and PT+CC baselines, which indicates the effectiveness of our methods. Comparison between PT and PT+CC proves that Contextualized Calibration is very effective in the zero-shot setting. The results of KPT-FR-RR-CC, which is the variant without any refinement, reveal the label noise is severe in the automatically constructed knowledgeable label words. 
The gap between KPT-FR-RR and KPT-FR-RR-CC is larger than the gap between PT+CC and PT, demonstrating the drastic difference in the prior probabilities of the knowledgeable label words as we hypothesized in ~\cref{sec: refine-CC}. Comparison between KPT, KPT-FR, KPT-FR-RR proves the effectiveness of the refinement methods.

For the analysis regarding each type of classification task, we observe that the performance boost compared to the baselines in topic classification is higher than sentiment classification, which we conjecture that topic classification requires more external knowledge than sentiment classification. While CC offers huge improvement (on average +13\%) over PT baseline, the incorporation of external knowledge further improves over PT+CC up to 11\% on DBPedia, and 6\% on AG's News and Yahoo. 
We also observe that the improvement brought by the refinement methods is more noticeable for topic classification tasks. By looking at the fraction of label words maintained after the refinement process (See appendix~\ref{app:remainwords}), we conjecture that the sentiment dictionary that we used in sentiment classification tasks contains little noise. Moreover, the improvement brought by the refinement process justifies the resilience of our methods to recover from noisy label words. 

\textbf{Few-shot.}\quad From Table~\ref{tab:experiment-few-shot}, we first find out that prompt-based methods win over fine-tuning by a dramatic margin
under nearly all situations. The gap enlarges as the shot becomes fewer. Comparing the baseline methods, the Soft Verbalizer (SOFT) generally wins over the Manual Verbalizer(PT) by a slight margin. 
However, automatic verbalizer (AUTO), although free of manual effort, lags behind the other verbalizers especially in a low-shot setting. The reason is obvious since the selection of label words among the vocabulary becomes inaccurate when labeled data is limited. 

When comparing \OURS with the baseline methods, we find \OURS or its variants consistently outperform all baseline methods. On average, 17.8\% , 10.3\%, and 7.4\% error rate reduction from the best baseline methods are achieved on 1, 5, and 10 shot experiments, respectively. Comparing within the variants of \OURS, we find that RR and LR are generally effective across shots on topic classification dataset, while in sentiment classification dataset,  \OURS works well without the refinements, which is consistent with our previous assumptions that the sentiment dictionary has little noise. Note that the KPT-RR variant does not utilize any unlabeled support set $\tilde{\mathcal{C}}$ since we do not conduct CC and FR by default in few-shot learning. This variant is still superior to the baseline methods in most cases. In terms of variance, we can see that \OURS enjoys smaller variances than baseline methods in most cases, demonstrating that the better coverage of label words stabilizes the training. 

For 20-shot experiments, we can see that  the gap between different methods narrows as the training data becomes sufficient. However, \OURS and its variants still win by a consistent margin
over the baseline methods. Surprisingly, with more training data, LR does not become more powerful as we may hypothesize. We conjecture that it is because all label words, even with some noise, \emph{can} serve as training objectives for prompt tuning. This perspective is similar to ~\citet{gao2020making} that using ``bad'' as a label word for the class ``positive'' can still preform classification although the performance degrades.

\section{Analysis}
\label{sec:analysis}
Ablation studies about our refinement methods have been shown in the previous section. In this section and Appendix~\ref{app:analysis}, we conduct more in-depth analyses on the proposed methods.





\subsection{Diversity of Top Predicted Words}
\begin{figure}[H]
    \centering
    \includegraphics[width=\linewidth]{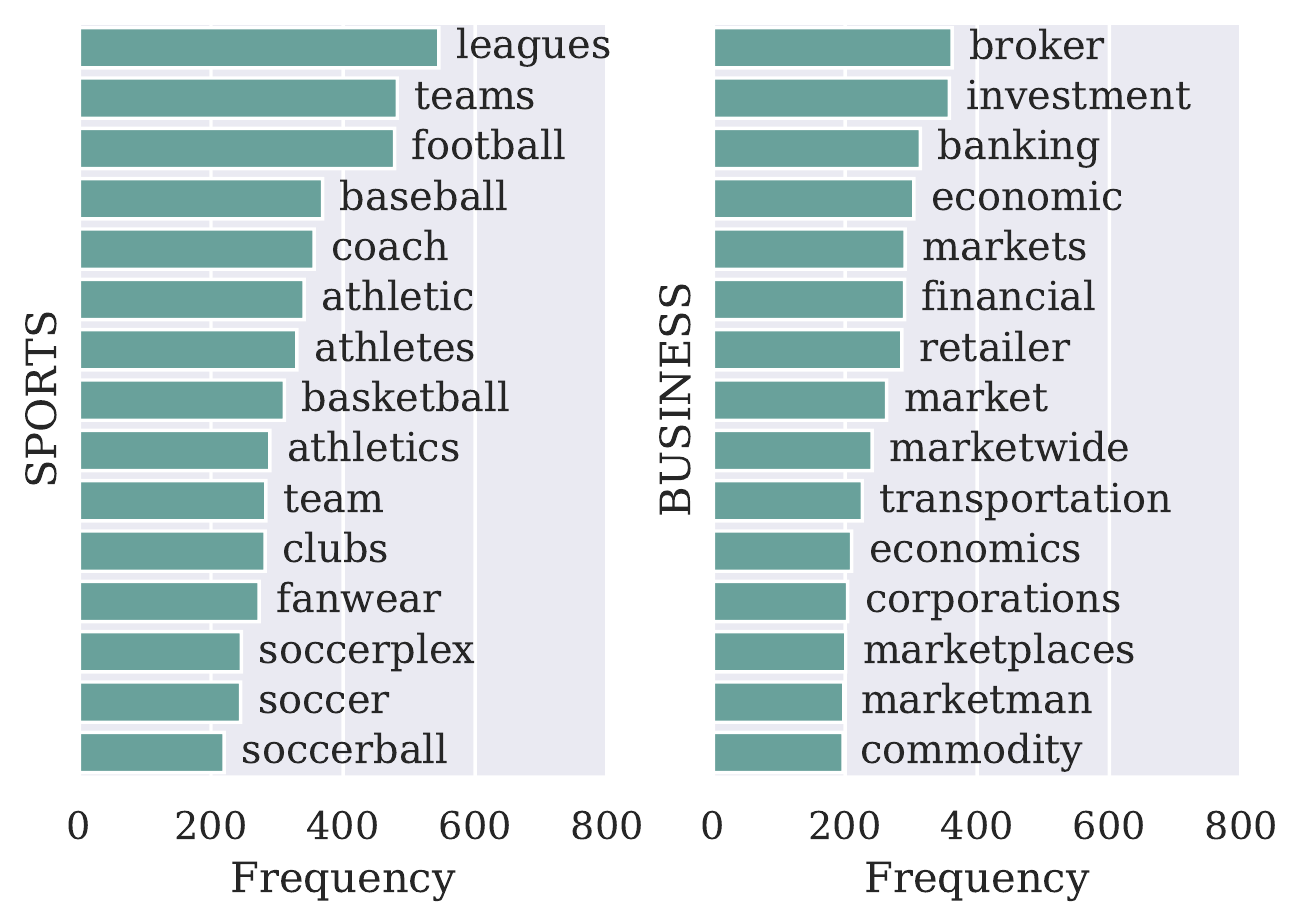}
    \caption{Frequent words appearing in the top-5 predictions. The results for two classes: \textsc{SPORTS} (left) and \textsc{BUSINESS} (right) are drawn.}
    \label{fig:violin}
\end{figure}

One advantage of KPT is that it can generate diverse label words across different granularities. To specifically quantify such diversity, we conduct a case study. For  the correctly predicted sentences of a class $y$, we count the frequency of label words $v\in \mathcal{V}_y$ appearing in the top-5 predictions for the $\texttt{[MASK]}$ position. Then we report the top-15 frequent label words in Figure~\ref{fig:violin}. Due to space limit, only the results of $\textsc{Sports}$ and $\textsc{Business}$ category of AG's News are shown.
As shown in Figure~\ref{fig:violin}, a diversity of label words, instead of mainly the original class names,  are predicted. And the predicted label words cover various aspects of the corresponding topic. For example, for the topic  \textsc{Sports}, the predicted ``leagues'', ``football'', and ``coach'' are related to it from different angles.

\subsection{Other Analyses}

In addition to the visualization, we study the influence of the support set's size on zero-shot text classification in Appendix~\ref{app:supportsize}. Then we justify that few-shot learning via labeled data eases the need for calibration and frequency-based refinement in Appendix~\ref{app:fewshot-cali}. We also demonstrate that our approach to handling the out-of-vocabulary (OOV) words is reasonable in Appendix~\ref{app:experiment-st}. Moreover, we take a closer look at the refinement process by analyzing the fraction of label words maintained during refinement in Appendix~\ref{app:remainwords}. Finally, we discuss the potential use of the proposed methods when knowledge bases resources are not readily available in Appendix~\ref{withoutkb}.



\section{Conclusion}
In this paper, we propose \OURS, which expands the verbalizer in prompt-tuning using the external KB. To better utilize the KB, we propose refinement methods for the knowledgeable verbalizer. The experiments show the potential of \OURS in both zero-shot settings and few-shot settings. For future work, there are  open questions related to our research for investigation: 
(1) Better approaches for combining KB and prompt-tuning in terms of template construction and verbalizer design.
(2) Incorporating external knowledge into prompt-tuning for other tasks such as text generation. 

\section*{Contributions}
Zhiyuan Liu, Huadong Wang and Shengding Hu proposed the idea and led the research. Shengding Hu designed the methods and conducted the experiments. Ning Ding and Shengding Hu wrote the abstract, introduction (Section~\ref{sec:intr}) and method (Section~\ref{sec:method}) part of the paper. Shengding Hu finished the other parts. Ning Ding, Huadong Wang and Zhiyuan Liu thoroughly revised the whole paper. Jingang Wang, Juanzi Li, Wei Wu and Maosong Sun proofread the paper and provided valuable comments.

\section*{Acknowledgements}
This work is supported by the National Key R\&D Program of China (No. 2020AAA0106502), Institute Guo Qiang at Tsinghua University, NExT++ project from the National Research Foundation, Prime Minister’s Office, Singapore under its IRC@Singapore Funding Initiative, and International Innovation Center of Tsinghua University, Shanghai, China.

\section*{Ethical Considerations}
This work proposes knowledgeable prompt tuning which uses external knowledge bases to construct the verbalizer. Users should be aware of the potential error in the external KBs, or even the injection of malicious words. 

\newpage

\bibliographystyle{acl_natbib}
\bibliography{acl.bib}
\newpage

\appendix

\section{Pilot Experiments}
\begin{table*}[tbp]
\begin{center}
\scalebox{0.80}{
\begin{tabular}{lllllll}
\toprule
Shot & Method  & AG's News   & DBPedia & Yahoo & Amazon & IMDB  \\
\midrule
\multirow{2}{*}{5} & SOFT &82.8 $\pm$ 2.7 \smallcolor{(84.3)} & 97.0 $\pm$ 0.6 \smallcolor{(97.2)} & 61.8 $\pm$ 1.8 \smallcolor{(63.1)} & 93.2 $\pm$ 1.6 \smallcolor{({94.2})} & 91.6 $\pm$ 3.4 \smallcolor{({93.9})} \\
& SOFT w.o. M & 63.4 $\pm$ 11.3 \smallcolor{(64.7)} & 82.1 $\pm$ 5.9 \smallcolor{(86.1)} & 24.5 $\pm$ 6.2 \smallcolor{(27.2)} & 79.2 $\pm$ 10.5 \smallcolor{(85.5)} & 83.6 $\pm$ 11.5 \smallcolor{(93.4)} \\
\midrule
\multirow{2}{*}{10} & SOFT &85.0 $\pm$ 2.8 \smallcolor{(86.7)} & 97.6 $\pm$ 0.4 \smallcolor{(97.8)} & 64.5 $\pm$ 2.2 \smallcolor{(65.0)} & 93.9 $\pm$ 1.7 \smallcolor{(93.9)} & 91.8 $\pm$ 2.6 \smallcolor{(93.0)} \\
& SOFT w.o. M & 77.4 $\pm$ 4.8 \smallcolor{(79.1)} & 94.9 $\pm$ 2.5 \smallcolor{(95.9)} & 42.6 $\pm$ 8.3 \smallcolor{(48.1)} & 92.9 $\pm$ 2.0 \smallcolor{(94.0)} & 88.7 $\pm$ 9.7 \smallcolor{(93.8)} \\
\bottomrule
\end{tabular}}
\end{center}
\caption{Pilot experiment on soft verbalizer justifies the need of human (expert) knowledge into the verbalizer. SOFT is the soft verbalizer with class name and SOFT w.o. M is the variant without the manual verbalizer.}
\label{tab:soft-pilot}
\end{table*}


\label{app:soft-pilot}
As pointed out by ~\citep{gao2020making}, manually defined verbalizer is competitive or even better than automatically searched/optimized verbalizers, which strengthens our motivation to improve over manual verbalizers by injecting more external human knowledge. To further illustrate the advantage of manual verbalizer, we conduct pilot experiments in soft verbalizer. Soft Verbalizer~\cite{hambardzumyan-etal-2021-warp} can be initialized with the predefined class name as the label words, which is adopted by us as a baseline in Table~\ref{tab:experiment-few-shot}. It can also be randomly initialized without the manually defined class names. We test the performance of Soft Verbalizer with and without the manually defined class name in 5 and 10 shot experiments. From Table~\ref{tab:soft-pilot}, we can see that the gaps between the variants are generally large. Therefore further improving the verbalizer with manually defined class name is a promising direction than the learned-from-scratch verbalizer without any human prior.

\section{A Theoretical Illustration of \OURS}
\label{theoretical}
In this section we provide a theoretical analysis of the whole framework used by \OURS. In prompt tuning, given a text $\mathbf{x}$, we wrap it into a template to form a wrapped sentence $\mathbf{x}_{\text{p}}$. We  then predict the probability of the label word $v$ using a PLM:
\begin{equation}
    p(\texttt{[M]}\!\!\!=\!v|\mathbf{x}) = P_\mathcal{M}(\texttt{[M]}\!\!\!=\!v|\mathbf{x}_{\text{p}}),
\end{equation}
where $\texttt{[M]}$ is short for $\texttt{[MASK]}$, denoting the label word's prediction at the masked position of the template. 

Then, if multiple label words are used to contribute to a single label, the predicted probability of the label is defined by marginalizing the probability of predicting all the label words, i.e., 
\begin{align}
\begin{split}
  p(\text{Y}\!=\!y|\mathbf{x}) \!=\!\! \sum_{v\in V_\mathcal{Y}} p(\text{Y}\!=\!y, \texttt{[M]}\!\!\!=\!v|\mathbf{x}).
\end{split}
\label{appequ_1}
\end{align}

Since the prediction of Y is independent of $\mathbf{x}$ given $v$, we can write Equation~\eqref{appequ_1} into
\begin{align}
\begin{split}
 &\sum_{v\in V_\mathcal{Y}} p(\text{Y}=y|\texttt{[M]}\!\!\!=\!v) p(\texttt{[M]}\!\!\!=\!v|\mathbf{x}) \\ 
 = & \sum_{v\in V_\mathcal{Y}}  p(\text{Y}=y|\texttt{[M]}\!\!\!=\!v) P_\mathcal{M}(\texttt{[M]}\!\!\!=\!v|\mathbf{x}_{\text{p}}).
 \end{split}
\label{appequ_2}
\end{align}

Using the Bayes Theorem and assuming a balanced classification problem,  Equation~\eqref{appequ_2} can be transformed into


\begin{align}\label{appequ_2}
 &\sum_{v\in V_\mathcal{Y}} \frac{p(\texttt{[M]}\!\!\!=\!v|\text{Y}=y)p(\text{Y}=y) }{p(\texttt{[M]}\!\!\!=\!v)} P_\mathcal{M}(\texttt{[M]}\!\!\!=\!v|\mathbf{x}_{\text{p}})\nonumber\\
 &\propto \!\! \sum_{v\in V_\mathcal{Y}}\! \frac{p(\texttt{[M]}\!\!\!=\!v|\text{Y}=y)}{p(\texttt{[M]}\!\!\!=\!v)} P_\mathcal{M}(\texttt{[M]}\!\!\!=\!v|\mathbf{x}_{\text{p}}). 
\end{align}

Now, the prediction probability of the label is composed of three parts.

(1) The first part $p(v|\text{Y}=y)$ is the probability of predicting the specific label word $v$ given the class label $y$. Intuitively, if a label word is relevant to label $y$, this term will be assigned a high probability. In \OURS, the \textbf{Relevance Refinement} estimate this probability using a quantized objective, i.e., if a relevance score exceeds the threshold $1$, it will be maintained, otherwise, it will be filtered. On the other hand, \textbf{Learnable Refinement} estimates this probability using continuous weights. 

(2) The second part is $p(\texttt{[M]}\!\!\!=\!v)$ in the denominator. This term is actually the prior probability of label words $v$, which is estimated by our \textbf{Contextualized Calibration}. Previous works also try to approach this term using a context-free manner~\cite{holtzman2021surface, zhao2021calibrate}.

(3) The last term  $P_\mathcal{M}(\texttt{[MASK]}\!\!\!=\!v|\mathbf{x}_{\text{p}})$ is the probability of the label words $v$ predicted by the PLM, which is the only component in most works such as Manual Verbalizers~\cite{schick2020exploiting}, yielding a sub-optimial solution compared to \OURS.

 Verbalizers with multiple label words for a class label can all be formalized into this framework once they use Equation~\eqref{appequ_1} as their backbone hypothesis. However, to the best of our knowledge, \OURS is the first to combine all of the three components to form a powerful verbalizer.

\section{Practical Issues of Refinement}
\label{App: filter}
In this section, we detail the refinement process by making 
some practical modifications to the methods in~\cref{sec:refine}.

\textbf{Frequency Refinement.} For Frequency Refinement, since the absolute value distribution of the contextualized prior probability may be different for each task,  determining a specific threshold of the contextualized prior probability may be tricky and elusive. We use a ranking-based threshold, i.e., we filter the label words that appear in the lower half of the contextualized prior probability.

\textbf{Relevance Refinement.} For Relevance Refinement, we observe that in the classification task with only a few classes, it's better to provide a stricter criterion to ensure that the relevance scores of a label word to \emph{any other} class is lower than the score to the belonging class, i.e., \emph{maximum} in the term of IDF-score is preferred.
To keep a unified criterion, we use a norm-based IDF-score.
\begin{equation}
    R^d(v)\! =\! r(v, f(v)) \large(\frac{|\mathcal{Y}|-1}{\sum_{y\in \mathcal{Y}, y\neq f(v)} r(v,y)^d}\large)^{1/d}
\end{equation}
where
\begin{equation}
    d = \frac{C}{|\mathcal{Y}|-2+\epsilon} +1, C>0.
\end{equation}
This criterion will approximate the maximum value in $\{r(v,y)|y\in |\mathcal{Y}|, y\neq f(v)\}$ in classification with only a few labels, and revert to the mean score in Equation~\eqref{equ: rr_tfidf} when conducting classification with many labels. We take $C=10$ (without trial and error) in the experiments. And $0 < \epsilon\ll 1$ is a small number to prevent numerical error.


\section{Further Analyses and Ablation Studies}
\label{app:analysis}
\begin{figure}
    \centering
    \includegraphics[width=\linewidth]{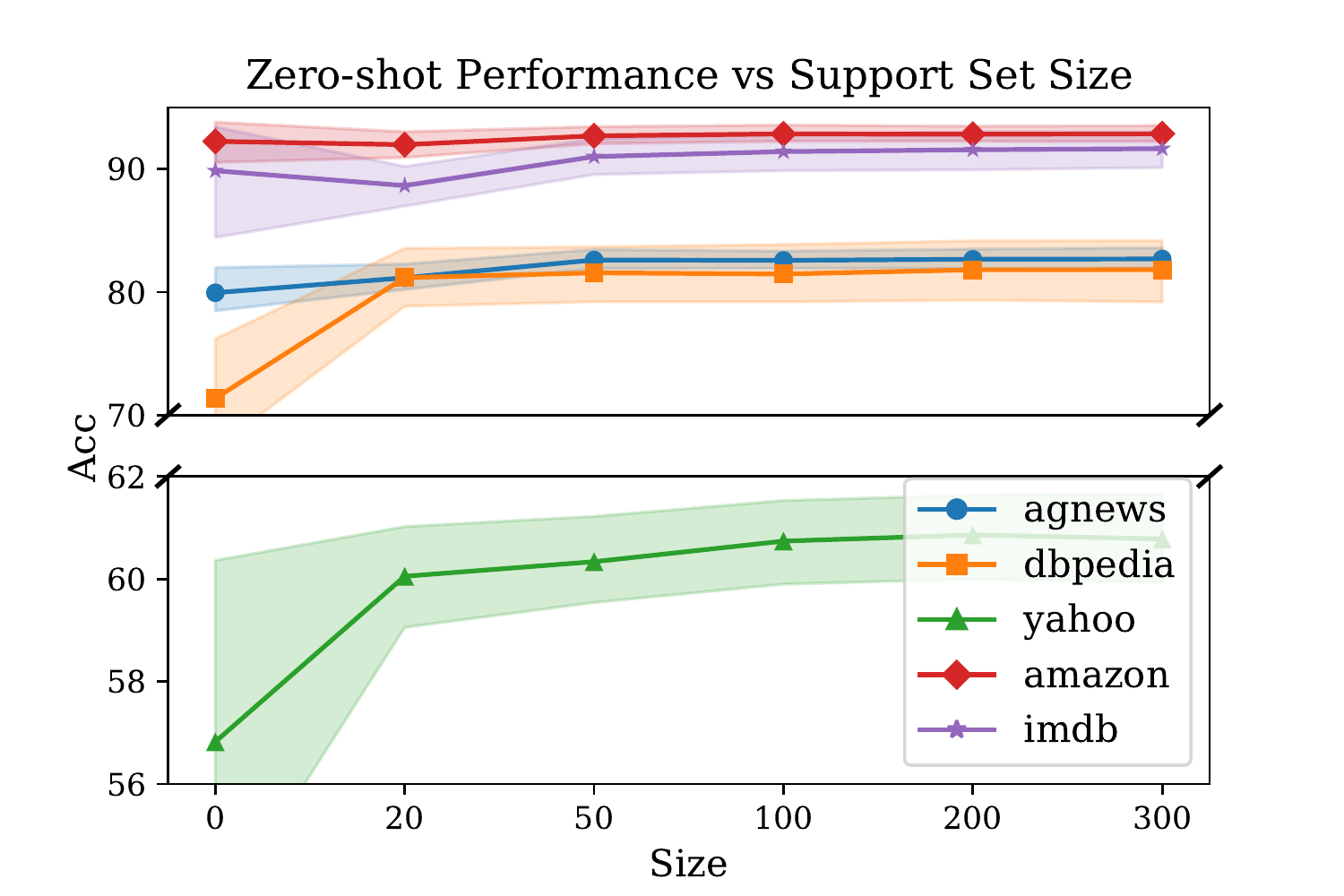}
    \caption{Support set size w.r.t. zero-shot performance. The points at size=0 are the performances of $\text{PMI}_\text{DC}$.}
    \label{fig:calistudy}
\end{figure}
\subsection{Calibration and Contextualized Calibration.}
\label{app:supportsize}
In fine-tuning, calibration has been studied under the topic of prediction confidence and out-of-distribution detection~\cite{kong-etal-2020-calibrated}. Recently, it got renascent attention in prompt learning~\cite{zhao2021calibrate, holtzman2021surface}.
In prompt learning, the PLM has a natural tendency to predict one word over another word regardless of the real sentence input. For example, GPT-3 prefers to predict ``positive'' over ``negative'' given ``N/A'' as the input sentence~\cite{zhao2021calibrate}. Therefore the calibration is crucial (see Table ~\ref{tab:experiment-zero-shot}) when no posterior optimization is conducted, i.e., in zero-shot learning. Existing methods such as $\text{PMI}_\text{DC}$ propose only using the empty template without filling the template with the instances in the corpus, for example, ``A \texttt{[Mask]} question :'', to produce the calibration logits. Our proposed Contextualized Calibration utilizes a limited amount of unlabeled support data to yield significantly better results. However, since we target the data-scarce scenario, we study in detail the amount of unlabeled data necessary to produce a satisfying calibration result. In Figure~\ref{fig:calistudy}, we draw the performance of \OURS - RR with different support set sizes $|\tilde{\mathcal{C}}|$. We also draw the performance of $\text{PMI}_\text{DC}$ on the $|\tilde{\mathcal{C}}|=0$ for comparison. 


From Figure~\ref{fig:calistudy}, we find that $|\tilde{\mathcal{C}}|\sim 50$ is enough to yield a satisfying calibration. Contextualized calibrate is more effective in classification with many classes, while calibrate without the context is effective in classification with few classes. 

In addition, we \emph{must point out} that if we have a set of sentences to classify in real-world scenarios, we can use these sentences themselves as the support set to conduct more accurate Contextualized Calibration.

\subsection{Supervised Data Ease the Need for Calibration.}
\label{app:fewshot-cali}
Although calibration is crucial for the zero-shot setting, we do not perform calibration for the few-shot setting because we assume that the posterior probability of the label words can be trained to the desired magnitude with only a few training instances. We also do not perform Frequency Refinement for few-shot learning due to the same assumption. To verify the assumption empirically, we add both Contextualized Calibration and Frequency Refinement to \OURS and test the performance under different settings. The results are shown in Table~\ref{tab:experiment-fewshot-cc-rf}. The performance comparison to \OURS without CC and FR in Table~\ref{tab:experiment-few-shot}  are shown using up arrows and down arrows. We can see that except in Yahoo, the improvement is not consistent for even negative, which supports our assumption that the need for calibration is greatly eased with the supervised input data.

\begin{table*}[!htbp]
\begin{center}
\scalebox{0.77}{
\begin{tabular}{lllllll}
\toprule
Shot & Method  & AG's News   & DBPedia & Yahoo & Amazon & IMDB  \\
\midrule
1 & \OURS + CC + FR  & 83.4 \bd $\pm$ 4.0 \smallcolor{(84.6)} & 94.0 \gd $\pm$ 2.0 \smallcolor{(95.7)} & 63.3 \gd $\pm$ 2.0 \smallcolor{(64.9)} & 93.2 \nc $\pm$ 1.2 \smallcolor{(94.0)} & 92.1 \bd $\pm$ 3.2 \smallcolor{(93.8)} \\
5 & \OURS + CC + FR  & 84.6 \bd $\pm$ 1.3 \smallcolor{(85.1)} & 97.3 \gd $\pm$ 0.3 \smallcolor{(97.4)} & 67.3 \gd $\pm$ 1.1 \smallcolor{(67.7)} & 94.0 \gd $\pm$ 1.2 \smallcolor{(94.7)} & 92.7 \nc $\pm$ 1.6 \smallcolor{(93.1)} \\
10 & \OURS + CC + FR  &  85.9 \bd $\pm$ 1.7 \smallcolor{(86.7)} & 98.1 \gd $\pm$ 0.2 \smallcolor{(98.2)} & 68.0 \nc $\pm$ 1.1 \smallcolor{(68.6)} & 93.3 \bd $\pm$ 1.8 \smallcolor{(93.7)} & 92.9 \nc $\pm$ 1.8 \smallcolor{(93.6)} \\
20 & \OURS+ CC + FR  & 87.3 \gd $\pm$ 0.8 \smallcolor{(87.6)} & 98.0 \bd $\pm$ 0.4 \smallcolor{(98.2)} & 69.1 \gd $\pm$ 0.7 \smallcolor{(69.5)} & 93.5 \bd $\pm$ 1.1 \smallcolor{(93.9)} & 93.1 \nc $\pm$ 1.3 \smallcolor{(93.5)} \\
\bottomrule
\end{tabular}}
\end{center}
\caption{Results of Contextualized Calibration and Frequency Refinement on few-shot experiments. The green up arrow \gd means the result is higher than \OURS in Table~\ref{tab:experiment-few-shot}, and the red down arrow \bd means the results is lower than \OURS in Table~\ref{tab:experiment-few-shot}.}
\label{tab:experiment-fewshot-cc-rf}
\end{table*}

\begin{table*}[!htbp]
\begin{center}
\scalebox{0.80}{
\begin{tabular}{lllllll}
\toprule
Shot & Method  & AG's News   & DBPedia & Yahoo & Amazon & IMDB  \\
\midrule
0 & \OURS + ST & 84.9 \gd $\pm$ 1.0 \smallcolor{(86.3)} & 81.0 \bd $\pm$ 4.3 \smallcolor{(85.2)} & 62.7 \gd $\pm$ 1.1 \smallcolor{(64.4)} & 92.8 \nc $\pm$ 1.2 \smallcolor{(94.7)} & 91.5 \bd $\pm$ 2.8 \smallcolor{(94.1)} \\
1 & \OURS + ST & 83.4 \bd $\pm$ 3.9 \smallcolor{(84.2)} & 94.0 \gd $\pm$ 1.8 \smallcolor{(95.8)} & 62.5 \bd $\pm$ 2.3 \smallcolor{(63.5)} & 93.3 \gd $\pm$ 1.4 \smallcolor{(94.1)} & 92.1 \bd $\pm$ 3.5 \smallcolor{(93.6)} \\
5 & \OURS + ST & 84.7 \bd $\pm$ 1.8 \smallcolor{(85.4)} & 97.1 \nc $\pm$ 0.5 \smallcolor{(97.2)} &66.8 \bd $\pm$ 1.0 \smallcolor{(67.3)} & 93.3 \bd $\pm$ 2.1 \smallcolor{(93.8)} & 93.1 \gd $\pm$ 1.4 \smallcolor{(93.3)} \\
10 & \OURS + ST & 86.3 \nc $\pm$ 1.5 \smallcolor{(86.8)} & 98.0 \nc $\pm$ 0.2 \smallcolor{(98.1)} & 67.6 \bd $\pm$ 0.9 \smallcolor{(67.9)} & 94.0  \gd $\pm$ 1.0 \smallcolor{(94.1)} & 92.7 \bd $\pm$ 1.8 \smallcolor{(93.6)} \\
20 & \OURS+ ST & 87.2 \bd $\pm$ 1.1 \smallcolor{(87.6)} & 97.9 \bd $\pm$ 0.4 \smallcolor{(98.1)} & 68.6 \bd $\pm$ 0.7 \smallcolor{(69.1)} & 93.5 \gd $\pm$ 1.8 \smallcolor{(94.0)} & 92.9 \bd $\pm$ 1.2 \smallcolor{(93.4)} \\
\bottomrule
\end{tabular}}
\end{center}
\caption{Results of restricting the expanded label word to be a single token in the PLM's vocabulary, where ST denotes ``single token''. The green up arrow \gd means the results is higher than \OURS in Table~\ref{tab:experiment-few-shot}, and the red down arrow \bd means the results is lower than \OURS in Table~\ref{tab:experiment-few-shot}.}
\label{tab:experiment-st}
\end{table*}

\subsection{How to Handle the OOV Label Words?}
\label{app:experiment-st}
Since the knowledgeable verbalizer is expanded using external resources which may not be tailored for the vocabulary of PLM. Thus, many label words are out-of-vocabulary (OOV) and are split into multiple tokens by the tokenizer. For these words, as mentioned in ~\cref{sec:refine}, we average the prediction probability of each token in the \emph{single} \texttt{[MASK]} position, which may not be very reasonable at the first glance. Therefore, we conduct an ablation study that whether forcing the label words to be a single token in the vocabulary of the PLM leads to better performance. The results under different shots are shown in Table~\ref{tab:experiment-st}. Surprisingly, 
making the single-token restriction does not yield stable improvement, instead, in many cases, the performance degrades by minor margins. Therefore we conclude that our method to handle OOV label words that are split by the tokenizer into multiple tokens is simple yet reasonable. More importantly, the label words expanded by the knowledge bases but not in the PLM's vocabulary \emph{can} serves as good label words in prompt tuning as well. 

\subsection{Visualization of the Refinement Process.}
\label{app:remainwords}
In this section, we report the number of label words that remained after Frequency Refinement and Relevance Refinement process. As we can see, these refinement methods remove a large fraction of label words while retaining the ones that are most informative. However, even the fewest number of remaining label words 
exceeds 100, which is far more than the number of label words in the previous works~\cite{schick2020automatically}. The broad coverage of label words contributes to the success of \OURS. 

\begin{figure}[!htbp]
    \centering
    \includegraphics[width=0.96\linewidth]{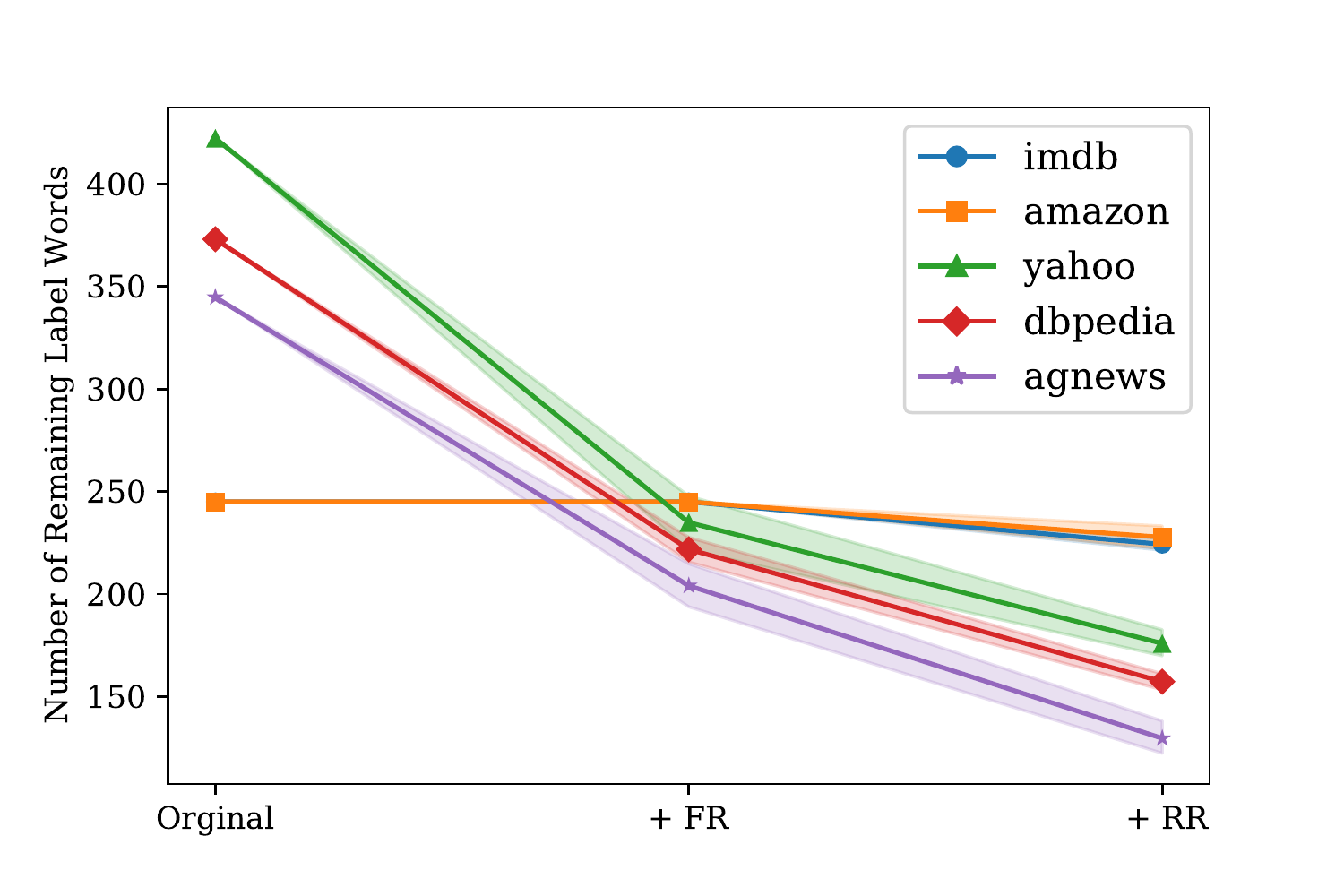}
    \caption{The number of remaining label words after Frequency Refinement and Relevance Refinement.}
    \label{fig:remainwords}
\end{figure}

\subsection{Potential Usage without External KB.}
\label{withoutkb}
Although KBs are ubiquitous in natural language processing, there are cases that no readily available KBs can be found for specific tasks. For these tasks, if we have enough unlabeled corpus, we can use the methods proposed by LOTClass~\cite{meng2020text} to mine potential label words from the corpus. More specifically, LOTClass~\cite{meng2020text} uses a self-supervised objective to train the PLM to extract the topic-related words from the whole unlabeled training corpus. Experiments that combine \OURS with LOTClass are beyond the scope of our work, but we believe the combination of the two can be very effective.

\section{Datasets and Templates}
We carry out experiments on three topic classification datasets: AG's News~\cite{zhang2015character}, DBPedia~\cite{lehmann2015dbpedia}, and Yahoo~\cite{zhang2015character}, and two sentiment classification datasets: IMDB~\cite{maas2011learning} and Amazon~\cite{mcauley2013hidden}. The statistics of the datasets are shown in Table~\ref{tab:dataset_stat}. 

\begin{table}[!htbp]
    \centering
    \scalebox{1}{
    \begin{tabular}{cccc}
\toprule
    Name&Type& \# Class& Test Size \\ 
\midrule
      AG's News   & Topic  & 4 & 7600 \\
       DBPedia  & Topic  & 14 & 70000 \\
       Yahoo & Topic & 10 & 60000 \\
       Amazon & Sentiment  & 2 & 10000 \\
       IMDB & Sentiment  & 2 & 25000 \\
     \bottomrule
    \end{tabular}
    }
    \caption{The statistics of each dataset.}
    \label{tab:dataset_stat}
\end{table}

\label{App:dataset-template}
 Due to the rich expert knowledge contained, the manual templates are proven to be competitive with or better than auto-generated templates~\cite{gao2020making} even though they are simpler to be constructed. Therefore we use manual templates in our experiments. Manual templates are also more applicable than auto-generated templates in the zero-shot setting. To mitigate the influence of different templates, we test \OURS under four manual templates that are either introduced by~\cite{schick2020exploiting} or tailored to fit the dataset for each experimental configuration. The templates for each dataset is listed below.

\textbf{AG's News}.\quad AG's News is a news' topic classification dataset. In this dataset, we follow PET~\cite{schick2020exploiting} to design the templates. However, their best performance pattern $T_1$(\textbf{x}) =  ``\texttt{[MASK]} news : \textbf{x}'' requires the \texttt{[MASK]}  token to be capitalized, which is not suitable for the label words in KB. And some of their templates are not informative and yield low performances. Therefore, we define four slightly changed templates:
\newtcolorbox{mybox}{colback=gray!6!white, colframe=gray!75!black}
\newtcolorbox{mybox2}{colback=gray!6!white, colframe=gray!75!black, width=0.53\textwidth}
\newtcolorbox{mybox3}{colback=gray!6!white, colframe=gray!75!black, width=0.50\textwidth}
\begin{mybox}
\vspace{-5mm}
\begin{align*}
     T_1(\mathbf{x})  &= \text{ A \texttt{[MASK]} news : }  \mathbf{x}  \\
     T_2(\mathbf{x})  &= \mathbf{x} \text{ This topic is about \texttt{[MASK]}. }\\
     T_3(\mathbf{x})   &= \text{ [ Category : \texttt{[MASK]} ] }  \mathbf{x}  \\
     T_4(\mathbf{x})   &=\text{ [ Topic : \texttt{[MASK]} ] } \mathbf{x} 
\end{align*}
\end{mybox}

\textbf{DBPedia}. \quad In a DBPedia sample, we are given a paragraph $\mathbf{b}$ paired with a title $\mathbf{a}$, in which the title is the subject of paragraph. The task is to determine the topic (or the type) of the subject. Different from other topic classifications, the paragraph can emphasize topics that are different from the title. For example, in a paragraph about an audio company, the main paragraph talks about music, albums, etc., but the correct label is ``company'' rather than ``music''. Therefore, we define the following templates:
\scalebox{0.9}{
\begin{mybox2}
\vspace{-5mm}
\begin{align*}
     T_1(\mathbf{a},\mathbf{b}) &=  \mathbf{a} \ \mathbf{b}  \  \tilde{\mathbf{a}}  \text{ is a \texttt{[MASK]} . }\\
      T_2(\mathbf{a},\mathbf{b}) &=  \mathbf{a} \   \mathbf{b}  \  \text{ In this sentence, } \  \tilde{\mathbf{a}}  \text{ is a \texttt{[MASK]} . } \\
   T_3(\mathbf{a},\mathbf{b}) &=   \mathbf{a}\ \mathbf{b} \  \text{The type of}\  \tilde{\mathbf{a}} \ \text{is}\  \texttt{[MASK]} .\\
   T_4(\mathbf{a},\mathbf{b}) &=\mathbf{a}\ \mathbf{b} \ \text{The category of}\  \tilde{\mathbf{a}} \ \text{is}\ \texttt{[MASK]} .
\end{align*}
\end{mybox2}
}
where $\tilde{\mathbf{a}}$ means removing the last punctuate in the title.

\textbf{Yahoo.} \quad Yahoo is a topic classification dataset about the questions raised in yahoo website~\cite{zhang2015character}. We use the same templates as AG's News, except that we change the word ``news'' into ``question'' in the $T_1(\mathbf{x})$: 
\begin{mybox}
\vspace{-5mm}
\begin{align*}
     T_1(\mathbf{x})  &= \text{ A \texttt{[MASK]} question : }  \mathbf{x}  \\
     T_2(\mathbf{x})  &= \mathbf{x} \text{ This topic is about \texttt{[MASK]}. }\\
     T_3(\mathbf{x})   &= \text{ [ Category : \texttt{[MASK]} ] }  \mathbf{x}  \\
     T_4(\mathbf{x})   &=\text{ [ Topic : \texttt{[MASK]} ] } \mathbf{x} 
\end{align*}
\end{mybox}

 \textbf{IMDB.}\quad IMDB is a sentiment classification dataset about movie reviews. Similar to the template defined in~\cite{schick2020exploiting} for sentiment classification, we define the following template:
\scalebox{0.96}{
\begin{mybox3}
\vspace{-5mm}
\begin{align*}
       T_1(\mathbf{x}) & = \text{ It was \texttt{[MASK]} . } \mathbf{x}\\
    T_2(\mathbf{x}) &  = \text{ Just \texttt{[MASK]} ! } \mathbf{x}\\
      T_3(\mathbf{x}) & = \mathbf{x} \text{ All in all, it was } \texttt{[MASK]} . \\
      T_4(\mathbf{x}) & = \mathbf{x} \text{ In summary, the film was } \texttt{[MASK]}.  
\end{align*}
\end{mybox3}
}

 \textbf{Amazon}. \quad Amazon is another sentiment classification dataset , we define the following template: 
\begin{mybox}
\vspace{-5mm}
\begin{align*}
       T_1(\mathbf{x}) & = \text{ It was \texttt{[MASK]} } .\  \mathbf{x}\\
    T_2(\mathbf{x}) &  = \text{ Just \texttt{[MASK]} ! } \mathbf{x}\\
      T_3(\mathbf{x}) & = \mathbf{x} \text{ All in all, it was  \texttt{[MASK]}} . \\
      T_4(\mathbf{x}) & = \mathbf{x} \text{ In summary, it was  \texttt{[MASK]}} ".  
\end{align*}
\end{mybox}
Since the test set of amazon is unnecessarily large for efficient testing, we randomly sample 10,000 samples from the 400,000 test samples to test, which is proven to have tiny influence on the performance in our pilot experiments.

\section{Experimental Settings}
\label{app:exp_settings}
We list the hyper-parameters in Table~\ref{tab:app_exp_settings}. Most of the hyper-parameters are the default parameters from Huggingface Transformers\footnote{\url{https://huggingface.co/transformers/}}.

\begin{table}[!htbp]
    \centering
    \begin{tabular}{p{0.2\textwidth}p{0.16\textwidth}p{0.05\textwidth}}
\toprule
      Hyper-parameter  & Dataset & Value \\
\midrule
   truncate length  & AG's News, DBPedia, Yahoo    &  128 \\
truncate length  & Amazon, Imdb    &  512 \\
   warmup steps & All & 0 \\
   learning rate & All & 3e-5 \\
   maximum epochs & All & 5\\
   adam epsilon & All & 1e-8 \\
\bottomrule
    \end{tabular}
    \caption{Hyper-parameter settings.}
    \label{tab:app_exp_settings}
\end{table}
For soft verbalizer, we use a learning rate of $3e-4$ to its soft label words' embeddings to encourage a faster convergence.

\end{document}